%% file: main.tex
\documentclass[sigconf]{acmart}
\usepackage{makecell}

\AtBeginDocument{%
  }

\acmConference[TSMO Workshop in conjunction with KDD]{Workshop on Two-sided Marketplace Optimization: Search, Pricing, Matching \& Growth in conjunction with KDD 2024}{2024}{Barcelona, Spain}

\setcopyright{none}
\renewcommand\footnotetextcopyrightpermission[1]{}
\begin{document}

\title{Predicting Potential Customer Support Needs and Optimizing Search Ranking in a Two-Sided Marketplace}

\author{Do-kyum Kim}
\affiliation{%
  \institution{Airbnb}
  \state{CA}
  \country{USA}}
\email{do-kyum.kim@airbnb.com}

\author{Han Zhao}
\affiliation{%
  \institution{Airbnb}
  \state{CA}
  \country{USA}}
\email{han.zhao@airbnb.com}

\author{Huiji Gao}
\affiliation{%
  \institution{Airbnb}
  \state{CA}
  \country{USA}}
\email{huiji.gao@airbnb.com}

\author{Liwei He}
\affiliation{%
  \institution{Airbnb}
  \state{CA}
  \country{USA}}
\email{liwei.he@airbnb.com}

\author{Malay Haldar}
\affiliation{%
  \institution{Airbnb}
  \state{CA}
  \country{USA}}
\email{malay.haldar@airbnb.com}

\author{Sanjeev Katariya}
\affiliation{%
  \institution{Airbnb}
  \state{CA}
  \country{USA}}
\email{sanjeev.katariya@airbnb.com}



\input{abstract}



\maketitle

\input{introduction}

\input{background}

\input{approach}

\input{experiment}

\input{discussion}

\begin{acks}
We thank our collaborators from the Relevance, Core Data Science and Community Support teams. Especially, support from Anna Matlin and Alex Deng was crucial in finding the right value of parameters. We also thank Airbnb's internal reviewers and anonymous reviewers for helpful comments.
\end{acks}


\bibliographystyle{ACM-Reference-Format}
\bibliography{reference}

\end{document}

%% file: abstract.tex
\begin{abstract}
Airbnb is an online marketplace that connects hosts and guests to unique stays and experiences. When guests stay at homes booked on Airbnb, there are a small fraction of stays that lead to support needed from Airbnb’s Customer Support (CS), which may cause inconvenience to guests and hosts and require Airbnb resources to resolve. In this work, we show that instances where CS support is needed may be predicted based on hosts and guests behavior. We build a model to predict the likelihood of CS support needs for each match of guest and host. The model score is incorporated into Airbnb’s search ranking algorithm as one of the many factors. The change promotes more reliable matches in search results and significantly reduces bookings that require CS support.
\end{abstract}

%% file: introduction.tex
\section{Introduction}
As of December 31, 2023, Airbnb had more than 7.7 million active listings from more than 5 million hosts worldwide \cite{web:2023stats}. While the platform invests heavily on its growth, it also strives to provide pleasant trip experience to both guests and hosts. In 2022, Airbnb launched AirCover that provides comprehensive protection for guests and hosts. For example, if a host cancels a reservation within 30 days of check-in, Airbnb provides support for finding a similar place, depending on availability at comparable pricing \cite{web:aircover:guest}.

While Airbnb provides Customer Support (CS) and Aircover, ideally, the need for CS support is minimized in the first place. That led to the key question of this paper: whether CS support needs may be predicted. If they are, some of them might be preventable before they happen and there is less need to contact CS. When we started this work, we had some evidence that they are. It is known that matching a new guest with a new host is more likely to require CS support as both parties are unfamiliar with how Airbnb works. Another example is same day bookings where more responsive hosts may result in less CS support needed.

Based on the evidence, we set out to build predictive models on whether a booking may result in CS support needs. From offline evaluation, we confirmed that our models are able to predict it to some extent. By incorporating the model in search ranking as one of the many considerations, we promote more reliable matching between guests and hosts, thereby preventing CS support needs before they happen.

%% file: background.tex
\section{Related work}
At Airbnb, search is the major interface where a guest is matched with a home given specific query parameters including location, dates and number of guests. During the match, a guest goes over a ranked list of homes and determines which one to book. Because of that, search ranking is one of the major levers for Airbnb to optimize business metrics. There are many different business metrics that can be optimized in search and recommendation systems. Ranking typically focuses on optimizing conversions. In Airbnb, we have been optimizing search ranking, either by directly predicting an uncancelled booking \cite{airbnb_kdd_2019}, or by adopting multi-task models to optimize multiple events through the guests’ search journey \cite{journey_ranker}.

Besides conversions, user satisfaction is another important business metric that is often optimized on search products for two sided marketplaces. User satisfaction is sometimes indirectly measured based on user engagement signals such as time between visits or retention rates. Alternatively, surveys are used to directly ask customers to rate their satisfaction. Search engines \cite{web:google17}\cite{web:google24} make continuous updates to their ranking algorithms to reduce the amount of low quality contents and fake news in their search results. Social media platforms \cite{web:meta} combine multiple user actions (e.g. clicks and likes) in their ranking algorithms, and they additionally apply integrity-related scores in the final stage to remove harmful contents. Video recommendation systems \cite{youtube-multitask}\cite{PLE} uses multi-task learning to optimize both user engagement and user satisfaction jointly to improve recommendation quality. E-commerce site \cite{AmazonMOO2023} uses multi-objective optimization algorithms to balance between objectives such as product quality and purchase likelihood.

%% file: approach.tex
\section{Approach}
We start by formulating a binary classification problem. To train a classifier, we construct a data set from past bookings on our platform. Lastly, the predicted likelihood from the classifier is incorporated into the ranking function in Airbnb homes search to help prevent CS support required.

\begin{figure}[h]
  \centering
  \includegraphics[width=\linewidth]{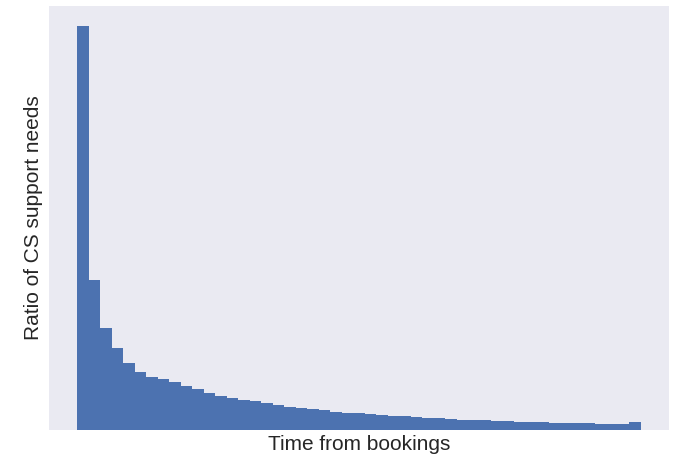}
  \caption{Time from bookings to CS support needs}
  \label{fig:ratio_vs_days}
\end{figure}

\subsection{Binary Classification}

We formulate a binary classification problem for predicting
\[p(\text{CS support needs} | \text{booking})\]. In words, our training examples are all the historical bookings, and positive examples are bookings that had CS support needs. Our models need to predict future outcomes that are significantly delayed; while CS support might be required anytime between booking and check outs (or even after that) — see Figure \ref{fig:ratio_vs_days} for a distribution —, we are trying to predict the probability at the time of bookings.

Thus, in constructing our data set, one of the key considerations is determining an attribution window, i.e. number of days to wait from a booking to determine a label. To increase coverage, we need to use a longer attribution window. On the other hand, using a longer attribution window prevents us from using the most recent data and it also has implications on running online A/B experiments. When we run an online experiment with a new search ranking algorithm, the experiment (and the new algorithm) is applied for a fixed time period but its outcome --- whether a booking made during the experiment led to CS support needs --- should be tracked for an extended period of time (at least longer than the attribution window).

In terms of input features, we consider all information available in our search ranking system since the model will be scored inside the system. When a guest searches for homes on Airbnb, they typically set filters such as destination, check in/out dates and number of guests. Given the search filters, we retrieve all the available homes that match the filters. For each available home, we will estimate the likelihood of CS support needs if the home is booked by the searcher with the search filters. Thus, we formulate features about the searcher, home and its hosts. We also found that search filters play important roles in predicting CS support needs. For example, same day bookings tend to require more CS support than other types of bookings.

\subsection{Maximizing Area under the ROC Curve}
The area under an ROC curve (AUC) is an evaluation metric used in many classification applications \cite{NIPS2003_6ef80bb2}. One interpretation of AUC is that, given pairs of positive and negative examples, it computes a ratio of pairs where a positive example is assigned a higher score by a model than a negative example is (see Lemma 1 in Cortes et al. \cite{NIPS2003_6ef80bb2}). Because of that, AUC is often used as a metric for ranking applications and we also used it as our main evaluation metric of models.

During model training, there are many approaches for directly maximizing AUC \cite{yang2022auc}. One simple method is, for a loss function, using an approximation of AUC based on sigmoid functions \cite{calders2007}. Specifically, an $i$-th example $x_i$ gets assigned a logit $f(x_i)$, where $f$ is a model. Next, between positive and negative examples, pairwise logit differences are computed and the differences are fed into sigmoids to form a loss function. Essentially, the loss function we optimize is equivalent to a cross entropy loss with only positive examples where each positive example is a pair of original positive and negative examples. Formally, it is computed as:
\[
- \sum_{x_i \in \text{positives}} \\  \sum_{x_j  \in \text{negatives}} \log \sigma(f(x_i) - f(x_j))
\], where $\sigma$ is a sigmoid function.

\subsection{Neural Networks}
There are different classes of models that can be used for classification and we choose to use neural networks \cite{Goodfellow-et-al-2016} for our application. One advantage of neural networks is that they can learn representation of categorical features during the course of training, which reduces efforts on manual feature engineering. It also guarantees parity with the current architecture and infrastructure as our main ranking models are also neural networks \cite{journey_ranker}.

To use neural networks, we normalized all continuous features so that each feature has zero mean and unit variance. For features with skewed distribution (e.g. number of past bookings), we also applied a logarithmic transformation before normalization. Each categorical feature is mapped to an embedding vector that is learned during the optimization process.

\subsection{Ranking Function}
Our current ranking function is a linear combination of predicted probabilities for many different events \cite{web:hc:search} including clicks on search results, sending booking requests and cancellations \cite{journey_ranker}. In the existing linear combination, we add one more term based on likelihood of CS support needs given a booking.

%% file: experiment.tex
\section{Experiments}
We optimized neural networks architectures and hyperparameters based on AUC metrics on a hold-out set. Another consideration was minimizing latency as these models will be computed for each search request at serving time. Finally, we choose one model and then run online A/B experiments to measure impacts on business metrics such as number of bookings and CS contacts.

\subsection{Data Set}
To train models, we collected historical bookings on Airbnb. Among the bookings that are past their attribution windows --- thus they have fixed labels ---, bookings made during the last three weeks were held out for evaluation, while those made in the year leading up to these three weeks were used for training.

\begin{table*}[t]
  \centering
  \caption{A few representative combinations of loss function, architectures and hyperparameters. `64-32' means that there are two hidden layers with 64 and 32 units respectively. The model from the first row is used in our online experiment.}
  \label{tab:hp}
  \begin{tabular}{lllrrr}
    \toprule
    Loss & Architecture & Hyperparameters & \# Multiplications & Validation AUC & Train AUC \\
    \midrule
    AUC & 128-64-32 with Relu & \makecell{batch size=5000, adam optimizer,\\l2 regularization=0.0005} & 36992 & 0.7321 & 0.7426 \\
    AUC & 64-32 with Relu & same as the 1st row & 15424 & 0.7314 & 0.7396 \\
    AUC & 512-256-64-32 with Relu & same as the 1st row & 256512 & 0.7326 & 0.7433 \\
    Cross entropy & same as the 1st row & same as the 1st row & 36992 & 0.7225 & 0.7317 \\
    AUC & 128-64-32 with Leaky Relu & same as the 1st row & 36992 & 0.7314 & 0.7421 \\
  \bottomrule
\end{tabular}
\end{table*}

\subsection{Hyperparameter Search}
We trained feed-forward neural networks with different depths and widths. For a loss function, either a cross entropy loss or the aforementioned approximation of AUC was used. We also varied other hyperparameters such as optimizers, regularizers and activation functions. Lastly, a learning rate scheduler was used to halve a learning rate when a plateau is reached on a validation set.

We list a few combinations and their AUC metrics in Table \ref{tab:hp}. Since AUC is a probability of correct ordering on pairs of positive and negative examples \cite{NIPS2003_6ef80bb2}, a random guesser would achieve an AUC of 0.5. Our models achieved AUC over 0.73, meaning CS support needs are predictable to some extent at the time of bookings. As expected, directly maximizing AUC outperformed minimizing cross entropy loss. Since positive examples --- bookings with CS support needs --- are rare, we used a large batch size. Increasing the number of hidden layers enhances the validation AUC, although with diminishing returns.

As our models need to serve online traffic, latency should be minimized. We assume that latency from model inference would be roughly proportional to the number of multiplications in a model since we are using CPUs to serve these models. In Table \ref{tab:hp}, we computed the number of multiplications in each architecture by using input dimension of 209 and number of parameters in the feed-forward neural networks. To balance between validation AUC and latency, we decided to use the model from the first row in Table \ref{tab:hp}.

\begin{figure}[h]
  \centering
  \includegraphics[width=\linewidth]{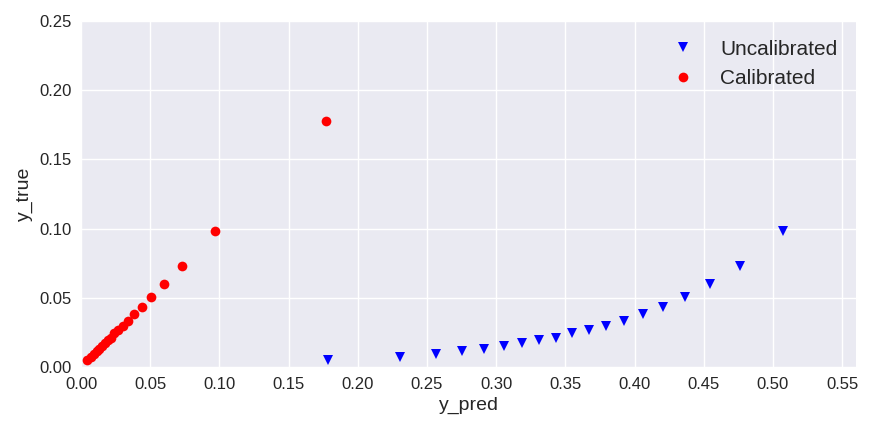}
  \caption{Calibration of model score using a Platt scaler. The line for calibrated score (red circle) is close to a diagonal line, which means the scaler provides well-calibrated probabilities.}
  \label{fig:calibration}
\end{figure}

\begin{figure}[h]
  \centering
  \includegraphics[width=\linewidth]{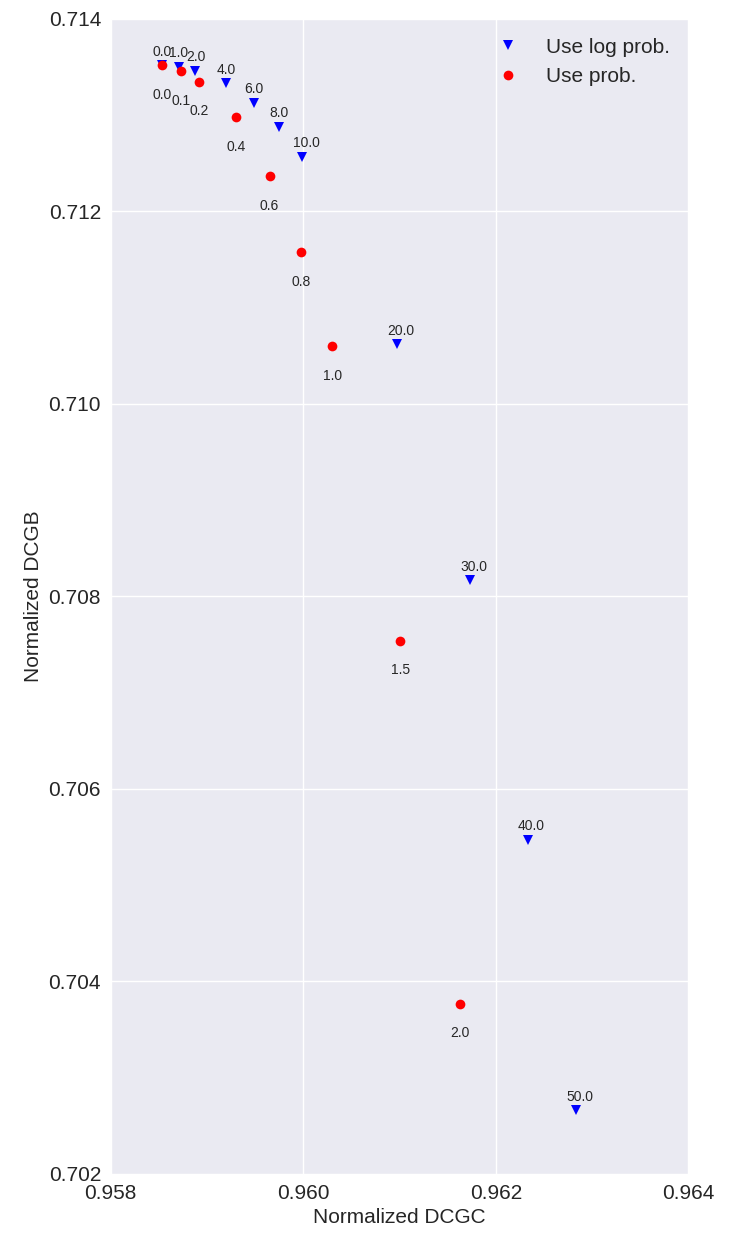}
  \caption{Trade off between normalized DCGB and normalized DCGC as a multiplier $\alpha$ is varied. We need to increase both of the metrics. For the details of metrics, please refer to section \ref{sec:tuning}.}
  \label{fig:tuning}
\end{figure}

\subsection{Offline Tuning of Ranking Function} \label{sec:tuning}
Since we train our models to maximize AUC, a resulting model does not provide calibrated probabilities. A Platt scaler \cite{platt} is used to calibrate model scores into conditional probabilities. The procedure is akin to training a logistic regressor with a single feature where the single feature is a logit from the underlying model $f(x)$ as:
\[
p(\text{CS support needs} | \text{booking}) = \frac{1}{1 + \exp(-w f(x) - b)}
\].
Figure \ref{fig:calibration} demonstrates the effectiveness of a Platt scaler for our application.

As mentioned in the previous section, our current ranking function is a linear combination of multiple conditional probabilities of different events in the conversion funnel. In the linear combination, we add one more term as:
\[
(\text{current ranking function}) + \alpha \log \big(1 - p(\text{CS support needs} | \text{booking}) \big)
\]

The multiplier $\alpha$ is determined by balancing two offline metrics on past search sessions. Each session consists of a query, a set of homes appeared in search results, and a label on which home is booked. In each session, a specific ranking function provides an ordering of the homes. For each ordering, we can compute a Discounted Cumulative Gain (DCG) \cite{ndcg_colt2013} with a booking label as:
\[
\text{DCGB} = \sum_{i=0}^{N}{\frac{r_i}{\log(2.4 + i)}}
\], where $N$ is a number of homes and $r_i$ is $1$ if the $i$-th home is booked. Similarly, we compute a DCG with conditional probabilities of a booking \textbf{without} CS support needs for each home as:
\[
\text{DCGC} = \sum_{i=0}^{N}{\frac{1 - p(\text{CS support needs} | i\text{-th home is booked})}{\log(2.4 + i)}}
\]
, where the conditional probabilities in numerator is from our model and Platt scaler.

On a data set of past search sessions from 7 days, we vary $\alpha$ and compute the two metrics DCGB and DCGC. The two metrics are normalized for each session so that they range between 0 and 1. The normalized metrics from a parameter sweep is presented in Figure \ref{fig:tuning}. As we increase $\alpha$, normalized DCGB declines, which means booked homes are ranked lower; but normalized DCGC improves, meaning that more reliable matchings are promoted in search results. It is also clear that using $\log$ probability provides a better trade-off than using the probability. We chose a few values for $\alpha$ to run online A/B experiments.

\subsection{Online A/B Experiments}

During our online experiments, we split searchers into multiple cohorts, and each cohort used a specific value of $\alpha$, which controls the strength of our model in the ranking function. For each cohort, we measured business metrics including number of bookings and also number of bookings with CS support needs. As expected, higher $\alpha$ reduced both number of bookings and number of bookings with CS support needs. We were able to find a right value of $\alpha$ that significantly reduces bookings with CS support needs while not negatively impacting overall booking conversions. Specifically, with the launched value of $\alpha$, we reduced bookings with CS support needs by 3.7\% and also lowered host cancellations by 2.7\% --- host cancellations often lead to CS contacts. These metrics align with our expectation that a better matching between guests and hosts can reduce CS support needs and benefit both sides in the marketplace.

%% file: discussion.tex
\section{Discussion}
We started from an assumption that some CS support needs might be predicted at the time of bookings. The assumption led us to build a binary classifier to predict if a booking — a match between a guest and a home (and the host) — would lead to CS support needs or not. From our offline analysis, we confirmed that CS support needs are predictable to some extent at the time of bookings.

By incorporating our prediction model in search ranking, we achieved a significant reduction in CS support needs without reducing overall conversions. Furthermore, we believe that reducing CS support needs can improve satisfaction of our guests and hosts, helping our guests to book more on Airbnb and also assisting our hosts in growing their business.

For future work, we can extend this approach to other types of user feedback to further improve trip experiences. To enhance model performance, we could experiment with more advanced model architectures and different loss functions. Another direction is exploring more ways to combine multiple model scores in a ranking function to better balance different business goals.